\documentclass{article}

\usepackage{times}
\usepackage{graphicx} 
\usepackage{subfigure} 

\usepackage{natbib}

\usepackage{algorithm}
\usepackage{algorithmic}

\usepackage{hyperref}

\usepackage[accepted]{icml2015}

\icmltitlerunning{Randomized Leverage-score Sketching}

\RequirePackage{amsthm,amsmath,amsfonts,amssymb}

\theoremstyle{definition}

\theoremstyle{remark}

\theoremstyle{plain}

\newtheorem{theo}{Theorem}[section]

\newtheorem{lem}{Lemma}[section]
\newtheorem{prop}{Proposition}[section]
\newtheorem{cor}{Corollary}[section]

\theoremstyle{definition} 

\newtheorem{nota}{Notation}[section]
\newtheorem{de}{Definition}[section]
\newtheorem{exa}{Example}[section]
\newtheorem{as}{Assumption}[section]
\newtheorem{alg}{Algorithm}[section]

\newcommand{\btheo}{\begin{theo}}
\newcommand{\bde}{\begin{de}}
\newcommand{\ble}{\begin{lem}}
\newcommand{\bpr}{\begin{prop}}
\newcommand{\bno}{\begin{nota}}
\newcommand{\bex}{\begin{exa}}
\newcommand{\bcor}{\begin{cor}}
\newcommand{\spro}{\begin{proof}}
\newcommand{\bas}{\begin{as}}
\newcommand{\balg}{\begin{alg}}

\newcommand{\etheo}{\end{theo}}
\newcommand{\ede}{\end{de}}
\newcommand{\ele}{\end{lem}}
\newcommand{\epr}{\end{prop}}
\newcommand{\eno}{\end{nota}}
\newcommand{\eex}{\end{exa}}
\newcommand{\ecor}{\end{cor}}
\newcommand{\fpro}{\end{proof}}
\newcommand{\eas}{\end{as}}
\newcommand{\ealg}{\end{alg}}

\theoremstyle{plain}

\newtheorem{theos}{Theorem}
\newtheorem{props}{Proposition}
\newtheorem{lems}{Lemma}
\newtheorem{cors}{Corollary}

\theoremstyle{definition}
\newtheorem{exas}{Example}
\newtheorem{algs}{Algorithm}
\newtheorem{asss}{Asumption}
\newtheorem{defns}{Definition}

\newcommand{\btheos}{\begin{theos}}
\newcommand{\etheos}{\end{theos}}
\newcommand{\bprops}{\begin{props}}
\newcommand{\eprops}{\end{props}}
\newcommand{\bdes}{\begin{defns}}
\newcommand{\edes}{\end{defns}}
\newcommand{\blems}{\begin{lems}}
\newcommand{\elems}{\end{lems}}
\newcommand{\bcors}{\begin{cors}}
\newcommand{\ecors}{\end{cors}}
\newcommand{\bexs}{\begin{exas}}
\newcommand{\eexs}{\end{exas}}
\newcommand{\balgs}{\begin{algs}}
\newcommand{\ealgs}{\end{algs}}
\newcommand{\bass}{\begin{asss}}
\newcommand{\eass}{\end{asss}}

\begin{document} 

\twocolumn[
\icmltitle{Statistical and Algorithmic Perspectives on Randomized \\ Sketching for Ordinary Least-Squares }

\icmlauthor{Garvesh Raskutti}{raskutti@stat.wisc.edu}
\icmladdress{University of Wisconsin, Madison,
            Department of Statitics, Department of Computer Science, Wisconsin Institute for Discovery, Madison, WI 53706 USA}
\icmlauthor{Michael W. Mahoney}{mmahoney@stat.berkeley.edu}
\icmladdress{University of California, Berkeley,
            ICSI, Department of Statistics, Berkeley, CA 94720 USA}

\icmlkeywords{boring formatting information, machine learning, ICML}

\vskip 0.3in
]

\begin{abstract} 
We consider statistical and algorithmic aspects of solving large-scale least-squares (LS) problems using randomized sketching algorithms.
Prior results show that, from an \emph{algorithmic perspective}, when using sketching matrices constructed from random projections and leverage-score sampling, if the number of samples $r$ much smaller than the original sample size $n$, then the worst-case (WC) error is the same as solving the original problem, up to a very small relative error. From a \emph{statistical perspective}, one typically considers the mean-squared error performance of randomized sketching algorithms, when data are generated according to a statistical linear model. In this paper, we provide a rigorous comparison of both perspectives leading to insights on how they differ. To do this, we first develop a framework for assessing, in a unified manner, algorithmic and statistical aspects of randomized sketching 
methods. We then consider the statistical prediction efficiency (PE) and the statistical residual efficiency (RE) of the sketched LS estimator; and 
we use our framework to provide upper bounds for several types of random projection and random sampling algorithms. Among other results, we show that the RE can be upper bounded when $r$ is much smaller than $n$, while the PE typically requires the number of samples $r$ to be substantially larger. Lower bounds developed in subsequent work show that our upper bounds on PE can not be improved.
\end{abstract} 

\section{Introduction}

Recent work in large-scale data analysis and Randomized Linear Algebra (RLA) has focused on developing so-called 
sketching algorithms: given a data set and an objective function of 
interest, construct a small ``sketch'' of the full data set, e.g., by using 
random sampling or random projection methods, and use that sketch as a 
surrogate to perform computations of interest for the full data 
set (see~\cite{Mah-mat-rev_BOOK} for a review).
Most effort in this area has adopted an \emph{algorithmic perspective}, 
whereby one shows that, when the sketches are constructed appropriately, 
one can obtain answers that are approximately as good as the exact answer
for the input data at hand, in less time than would be required to compute 
an exact answer for the data at hand.
From a \emph{statistical perspective}, however, one is often more interested in how well a procedure
performs relative to an hypothesized model than how well it performs on the 
particular data set at hand. Thus an important question to consider is whether
the insights from the algorithmic perspective of sketching carry over to the statistical 
setting. 
To address this, in this paper, we develop a unified approach that considers both the 
 \emph{statistical perspective} and \emph{algorithmic perspective} on 
recently-developed randomized sketching algorithms in RLA, and we provide bounds on 
two statistical objectives for several types of random projection 
and random sampling sketching algorithms.

\subsection{Overview of the problem}

The problem we consider in this paper is the ordinary least-squares (LS or 
OLS) problem:
given as input a matrix $X \in \mathbb{R}^{n \times p}$ of observed features 
or covariates and a vector $Y \in \mathbb{R}^n$ of observed responses,  
return as output a vector $\beta_{OLS}$ that solves the following 
optimization problem:

\begin{equation}
\label{NoiseLinMod}
\beta_{OLS} = \arg \min_{\beta\in\mathbb{R}^p}\|Y - X \beta\|_2^2  .
\end{equation}

\noindent
We will assume that $n$ and $p$ are both very large, with $n \gg p$, and 
for simplicity we will assume $\mbox{rank}(X) = p$, e.g., to ensure a unique 
full-dimensional solution. 
The LS solution, $\beta_{OLS} = (X^T X)^{-1} X^T Y$, has a number of 
well-known desirable statistical properties~\cite{ChatterjeeHadi88}; and 
it is also well-known that the running time or computational complexity for 
this problem is $O(n p^2)$~\cite{GVL96}.%
\footnote{That is, $O(n p^2)$ time suffices to compute the LS solution 
from Problem~(\ref{NoiseLinMod}) for arbitrary or worst-case input, with, 
e.g., the Cholesky Decomposition on the normal equations, with a QR 
decomposition, or with the Singular Value Decomposition~\cite{GVL96}.}
For many modern applications, however, $n$ may be on the order of 
$10^6-10^9$ and $p$ may be on the order of $10^3-10^4$, and thus computing 
the exact LS solution with traditional $O(n p^2)$ methods can be 
computationally challenging. 
This, coupled with the observation that approximate answers often suffice 
for downstream applications, has led to a large body of work on developing 
fast approximation algorithms to the LS problem~\cite{Mah-mat-rev_BOOK}.

One very popular approach to reducing computation is to perform LS on a 
carefully-constructed ``sketch'' of the full data set. 
That is, rather than computing a LS estimator from 
Problem~(\ref{NoiseLinMod}) from the full data $(X,Y)$, generate ``sketched 
data'' $(SX, SY)$ where $S \in \mathbb{R}^{r \times n}$, with $r \ll n$, is a 
``sketching matrix,'' and then compute a LS estimator from the following 
sketched problem: 

\begin{equation}
\label{NoiseLinModSketched}
\beta_S = \arg \min_{\beta \in \mathbb{R}^p}\|SY - S X \beta\|_2^2. 
\end{equation}

\noindent
Once the sketching operation has been performed, the additional computational 
complexity of $\beta_S$ is $O(r p^2)$, i.e., simply call a 
traditional LS solver on the sketched problem. 
Thus, when using a sketching algorithm, two criteria are important:
first, ensure the accuracy of the sketched LS estimator is comparable to,
e.g., not much worse than, the performance of the original LS estimator; and 
second, ensure that computing and applying the sketching matrix $S$ is not 
too computationally intensive, e.g., that it is much faster than solving the 
original problem exactly.

\subsection{Prior results}

Random sampling and random projections provide two approaches to construct 
sketching matrices $S$ that satisfy both of these criteria and that have 
received attention recently in the computer science community. In terms of 
running time guarantees, the running time bottleneck for random 
projection algorithms for the LS problem is the application of the 
projection to the input data, i.e., actually performing the matrix-matrix
multiplication to implement the projection and compute the sketch. 
By using fast Hadamard-based random projections, however,
Drineas et al.~\cite{DrinMuthuMahSarlos11} developed a random 
projection algorithm that runs on arbitrary or worst-case input 
in $o(np^2)$ time.
(See~\cite{DrinMuthuMahSarlos11} for a precise statement of the running 
time.)
As for random sampling, Drineas et al.~\cite{DMM06,DMMW12_JMLR} have shown that 
if the random sampling is performed with respect to nonuniform importance
sampling probabilities that depend on the \emph{empirical statistical 
leverage scores} of the input matrix $X$, i.e., the diagonal entries of the 
\emph{hat matrix} $H = X(X^T X)^{-1} X^T$, then one obtains a random 
sampling algorithm that achieves much better results for arbitrary or 
worst-case input.%

Leverage scores have a long history in robust statistics and experimental design. In the robust statistics community, samples with high leverage scores are typically flagged as potential 
outliers (see, e.g.,~\cite{ChatterjeeHadi86,ChatterjeeHadi88, Hampel86, HW78, Huber81}). In the experimental design community, samples with high leverage have been shown to improve overall efficiency, provided that the underlying statistical model is accurate (see, e.g.,~\cite{Royall70, Zaslavsky08}). This should be contrasted with their use in theoretical computer science. From the algorithmic perspective of worst-case analysis, that was adopted by Drineas et al.~\cite{DrinMuthuMahSarlos11} and Drineas et al.~\cite{DMMW12_JMLR}, samples with high leverage tend to contain the most important information for subsampling/sketching.  Thus it is beneficial for worst-case analysis to bias the random sample to 
include samples with large leverage scores or to rotate with a random projection to a random basis where the leverage scores are approximately uniformized. 

The running-time bottleneck for this leverage-based random sampling 
algorithm is the computation of the leverage scores of the input data; and
the obvious well-known algorithm for this involves $O(np^2)$ time to perform a QR 
decomposition to compute an orthogonal basis for $X$~\cite{GVL96}. 
By using fast Hadamard-based random projections, however, Drineas 
et al.~\cite{DMMW12_JMLR} showed that one can compute approximate QR 
decompositions and thus approximate leverage scores in $o(np^2)$ time; and
(based on previous work~\cite{DMM06}) this immediately implies a 
leverage-based random sampling algorithm that runs on arbitrary or 
worst-case input in $o(np^2)$ time~\cite{DMMW12_JMLR}. 
Readers interested in the practical performance of these randomized 
algorithms should consult \textsc{Bendenpik}~\cite{AMT10} or 
\textsc{LSRN}~\cite{MSM14_SISC}.

In terms of accuracy guarantees, both 
Drineas et al.~\cite{DrinMuthuMahSarlos11} and 
Drineas et al.~\cite{DMMW12_JMLR} prove that their respective random 
projection and leverage-based random sampling LS sketching algorithms each 
achieve the following worst-case (WC) error guarantee:
for any arbitrary $(X, Y)$,
\begin{equation}
\label{eqn:ErrorWCE}
\|Y - X \beta_S \|_2^2 \leq (1+ \kappa)\|Y - X \beta_{OLS} \|_2^2,
\end{equation} 
with high probability for some pre-specified error parameter $\kappa \in (0,1)$.
This $1+ \kappa$ relative-error guarantee%
\footnote{The nonstandard parameter $\kappa$ is used here for the error 
parameter since $\epsilon$ is used below to refer to the noise or error 
process.}
is extremely strong, and it is applicable to arbitrary or worst-case input.
That is, whereas in statistics one typically assumes a model, e.g., a 
standard linear model on $Y$,
\begin{equation}
\label{EqnLinModel}
Y = X \beta + \epsilon,
\end{equation}
where $\beta \in \mathbb{R}^p$ is the true parameter and 
$\epsilon \in \mathbb{R}^n$ is a standardized noise vector, with 
$\mathbb{E}[\epsilon]=0$ and $\mathbb{E}[\epsilon\epsilon^T]=I_{n \times n}$, 
in Drineas et al.~\cite{DrinMuthuMahSarlos11} 
and Drineas et al.~\cite{DMMW12_JMLR}
no statistical model is assumed on $X$ and $Y$, and 
thus the running time and quality-of-approximation 
bounds apply to any arbitrary $(X,Y)$ input data. 

\subsection{Our approach and main results}

In this paper, we address the following fundamental 
questions. 
First, under a standard linear model, e.g., as given in 
Eqn.~(\ref{EqnLinModel}), what properties of a sketching matrix $S$ are 
sufficient to ensure low statistical error, e.g., mean-squared error?
Second, how do existing random projection algorithms and leverage-based 
random sampling algorithms perform by this statistical measure? 
Third, how does this relate to the properties of a sketching matrix $S$ that
are sufficient to ensure low worst-case error, e.g., of the form of
Eqn.~(\ref{eqn:ErrorWCE}), as has been established 
previously~\cite{DrinMuthuMahSarlos11,DMMW12_JMLR,Mah-mat-rev_BOOK}?
We address these related questions in a number of steps. 

In Section~\ref{SecFramework}, we will present a framework for evaluating
the algorithmic and statistical properties of randomized 
sketching methods in a unified manner; and we will show that providing 
WC error bounds of the form of Eqn.~(\ref{eqn:ErrorWCE}) and 
providing bounds on two related statistical objectives boil down to 
controlling different structural properties of how the sketching matrix $S$ 
interacts with the left singular subspace of the design matrix.
In particular, we will consider the oblique projection matrix,
$\Pi_S^U = U (SU)^{\dagger} S$, where $(\cdot)^\dagger$ denotes the 
Moore-Penrose pseudo-inverse of a matrix and $U$ is the left singular matrix 
of $X$. 
This framework will allow us to draw a comparison between the WC error
and two related statistical efficiency criteria, the statistical 
prediction efficiency (PE) (which is based on the prediction error 
$\mathbb{E}[\|X(\widehat{\beta} - \beta)\|_2^2]$ and which is given in 
Eqn.~(\ref{DefnSPE}) below) and the statistical residual efficiency (RE)
(which is based on residual error $\mathbb{E}[\|Y - X \widehat{\beta}\|_2^2]$ and 
which is given in Eqn.~(\ref{DefnSRE}) below); and 
it will allow us to provide sufficient conditions that any sketching matrix 
$S$ must satisfy in order to achieve performance guarantees for these two 
statistical objectives. 

In Section~\ref{SecMainResults}, we will present our main theoretical 
results, which consist of bounds for these two statistical quantities for 
variants of random sampling and random projection sketching algorithms.
In particular, we provide upper bounds on the PE and RE (as well as the 
worst-case WC) for four sketching schemes: 
(1) an approximate leverage-based random sampling algorithm, as is 
analyzed by Drineas et al.~\cite{DMMW12_JMLR}; 
(2) a variant of leverage-based random sampling, where the random samples
are \emph{not} re-scaled prior to their inclusion in the sketch, as is 
considered by Ma et al.~\cite{MMY15_JMLR}; 
(3) a vanilla random projection algorithm, where $S$ is a random matrix 
containing i.i.d. Gaussian or Rademacher random variables, as is popular in 
statistics and scientific computing; and 
(4) a random projection algorithm, where $S$ is a random 
Hadamard-based random projection, as analyzed in~\cite{BoutsGitt12}. 
For sketching schemes (1), (3), and (4), our upper bounds for each of the two 
measures of statistical efficiency are identical up to constants; and
they show that the RE scales as $1+ \frac{p}{r}$, while the PE scales as 
$\frac{n}{r}$. 
In particular, this means that it is possible to obtain good bounds for the 
RE when $p \lesssim r \ll n$ (in a manner similar to the sampling 
complexity of the WC bounds); but in order to obtain even near-constant 
bounds for PE, $r$ must be at least of constant order compared to $n$.
For the sketching scheme (2), we show, on the other hand, that under the 
(strong) assumption that there are $k$ ``large'' leverage scores and the 
remaining $n-k$ are ``small,'' then the WC scales as $1+ \frac{p}{r}$, the 
RE scales as $1+ \frac{pk}{rn}$, and the PE scales as $\frac{k}{r}$.
That is, sharper bounds are possible for leverage-score sampling without 
re-scaling in the statistical setting, but much stronger assumptions are 
needed on the input~data. 
We also present a lower bound developed in subsequent work by Pilanci 
and Waniwright~\cite{PilanciWainwright} which shows that under general conditions
on $S$, the upper bound of $\frac{n}{r}$ for PE can not be improved. Hence
our upper bounds in Section~\ref{SecMainResults} on PE can not be improved.

In Section~\ref{SecDiscussion}, we will provide a brief discussion and 
conclusion.
For space reasons, we do not include in this conference version the proofs of our main results or our empirical results that support our theoretical findings; but they are included in the technical report version of this paper~\cite{RaskuttiMahoney}.

\subsection{Additional related work}

Very recently, Ma et al.~\cite{MMY15_JMLR} considered 
statistical aspects of leverage-based sampling algorithms (called 
\emph{algorithmic leveraging} in~\cite{MMY15_JMLR}).
Assuming a standard linear model on $Y$ of the form of 
Eqn.~(\ref{EqnLinModel}), the authors developed first-order Taylor 
approximations to the statistical RE of different 
estimators computed with leverage-based sampling algorithms, and they 
verified the quality of those approximations with computations on real and 
synthetic data. 
Taken as a whole, their results suggest that, if one is interested in the
statistical performance of these randomized sketching algorithms, then there 
are nontrivial trade-offs that are not taken into account by standard
WC analysis.
Their approach, however, does not immediately apply to random projections or other more 
general sketching matrices. Further, the realm of applicability of the first-order Taylor approximation was not precisely quantified, and they left open the question of structural characterizations of random sketching matrices that were sufficient to ensure good statistical properties on the sketched data. We address these issues in this paper.

Subsequent work by Pilanci and Wainwright~\cite{PilanciWainwright} also considers a statistical perspective of sketching. Amongst other results, they develop a lower bound which confirms that using a single randomized sketching matrix $S$ can not achieve a better PE than $\frac{n}{r}$. This lower bound complements the upper bounds developed in this paper. Their main focus is to use this insight to develop an iterative sketching scheme which yields bounds on the SPE when an $ r\times n$ sketch is applied repeatedly.

\section{General framework and structural results}
\label{SecFramework}

In this section, we develop a framework that allows us to view the
algorithmic and statistical perspectives on LS problems from a 
common perspective. We then use this framework to show that existing worst-case bounds as well as our novel statistical bounds for the mean-squared errors can be expressed in terms of different structural conditions on how the sketching matrix $S$ interacts with the data $(X,Y)$. 

\subsection{A statistical-algorithmic framework}

Recall that we are given as input a data set, 
$(X, Y) \in \mathbb{R}^{n\times p} \times \mathbb{R}^n$, and the objective 
function of interest is the standard LS objective,
as given in Eqn.~(\ref{NoiseLinMod}).
Since we are assuming, without loss of generality, that 
$\mbox{rank}(X)=p$, we have that 
\begin{equation}
\label{eqn:beta_opt_full}
\beta_{OLS} = X^{\dagger}Y = (X^T X)^{-1}X^T Y, 
\end{equation}
where $(\cdot)^{\dagger}$ denotes the Moore-Penrose pseudo-inverse of a matrix. 

To present our framework and objectives, let $S \in \mathbb{R}^{r \times n}$ 
denote an \emph{arbitrary} sketching matrix. 
That is, although we will be most interested in sketches constructed from 
random sampling or random projection operations, for now we let $S$ be 
\emph{any} $r \times n$ matrix.
Then, we are interested in analyzing the performance of objectives 
characterizing the quality of a ``sketched'' LS objective, as given in 
Eqn~(\ref{NoiseLinModSketched}), where again we are interested in solutions of the form 
\begin{equation}
\label{eqn:beta_opt_sketched}
\beta_S=(SX)^{\dagger}SY . 
\end{equation}
(We emphasize that this does \emph{not} in general equal 
$((SX)^T SX)^{-1}(SX)^T SY$, since the inverse will \emph{not} exist if the 
sketching process does not preserve rank.)
Our goal here is to compare the performance of $\beta_S$ to
$\beta_{OLS}$.
We will do so by considering three related performance criteria, two of a statistical flavor, and one of a more algorithmic or worst-case flavor.

From a statistical perspective, it is common to assume a standard linear 
model on $Y$, 
\begin{equation*}
Y = X \beta + \epsilon,
\end{equation*}
where we remind the reader that $\beta \in \mathbb{R}^p$ is the true 
parameter and $\epsilon \in \mathbb{R}^n$ is a standardized noise 
vector, with $\mathbb{E}[\epsilon]=0$ and 
$\mathbb{E}[\epsilon\epsilon^T]=I_{n \times n}$. 
From this statistical perspective, we will consider the following two~criteria.
\begin{itemize}
\item
The first statistical criterion we consider is the
\emph{prediction efficiency} (PE), defined as follows:
\begin{equation}
\label{DefnSPE}
C_{PE}(S) = \frac{\mathbb{E}[\|X (\beta - \beta_S)\|_2^2]}{\mathbb{E}[\|X (\beta - \beta_{OLS})\|_2^2]}  , 
\end{equation}
where the expectation $\mathbb{E}[\cdot]$ is taken over the random noise 
$\epsilon$. 
\item
The second statistical criterion we consider is the 
\emph{residual efficiency} (RE), defined as follows:
\begin{equation}
\label{DefnSRE}
C_{RE}(S) = \frac{\mathbb{E}[\|Y - X \beta_S\|_2^2]}{\mathbb{E}[\|Y - X \beta_{OLS}\|_2^2]}  , 
\end{equation}
where, again, the expectation $\mathbb{E}[\cdot]$ is taken over the random 
noise $\epsilon$.
\end{itemize}
Recall that the standard relative statistical efficiency for two estimators 
$\beta_1$ and $\beta_2$ is defined as 
$\mbox{eff}(\beta_1,\beta_2)=\frac{\mbox{Var}(\beta_1)}{\mbox{Var}(\beta_2)}$, 
where $\mbox{Var}(\cdot)$ denotes the variance of the 
estimator (see, e.g.,~\cite{Lehmann98}). 
For the PE, we have replaced the variance of each estimator by the 
mean-squared prediction error. 
For the RE, we use the term residual since for any estimator 
$\widehat{\beta}$, $Y - X \widehat{\beta}$ are the residuals for estimating 
$Y$. 

From an algorithmic perspective, there is no noise process $\epsilon$.
Instead, $X$ and $Y$ are arbitrary, and $\beta$ is simply computed 
from Eqn~(\ref{eqn:beta_opt_full}).
To draw a parallel with the usual statistical generative process, however, 
and to understand better the relationship between various objectives, 
consider ``defining'' $Y$ in terms of $X$ by the following ``linear model'':
\begin{equation*}
Y = X \beta + \epsilon,
\end{equation*}
where $\beta \in \mathbb{R}^p$ and $\epsilon \in \mathbb{R}^n$.  
Importantly, $\beta$ and $\epsilon$ here represent different quantities than 
in the usual statistical setting. 
Rather than $\epsilon$ representing a noise process and $\beta$ 
representing a ``true parameter'' that is observed through a noisy $Y$, 
here in the algorithmic setting, we will take advantage of the 
rank-nullity theorem in linear algebra to relate $X$ and $Y$.%
\footnote{The rank-nullity theorem asserts that given any 
matrix $X \in \mathbb{R}^{n \times p}$ and vector $Y \in \mathbb{R}^n$, 
there exists a unique decomposition $Y = X \beta + \epsilon$, where $\beta$ 
is the projection of $Y$ on to the range space of $X^T$ and 
$\epsilon = Y-X\beta$ lies in the null-space of $X^T$~\cite{Meyer00}.}   
To define a ``worst case model'' $ Y = X \beta + \epsilon$ for the 
algorithmic setting, one can view the ``noise'' process $\epsilon$ to 
consist of any vector that lies in the null-space of $X^T$.
Then, since the choice of $\beta \in \mathbb{R}^p$ is arbitrary, one can 
construct any arbitrary or worst-case input data $Y$.
From this algorithmic case, we will consider the following~criterion.
\begin{itemize}
\item
The algorithmic criterion we consider is the \emph{worst-case} (WC) error, 
defined as follows:
\begin{equation}
\label{DefnWCE}
C_{WC}(S) = \sup_{Y} \frac{\|Y - X \beta_S\|_2^2}{\|Y - X \beta_{OLS}\|_2^2}.
\end{equation}
\end{itemize}
This criterion is worst-case since we take a supremum $Y$, 
and it is the performance criterion that is analyzed in 
Drineas et al.~\cite{DrinMuthuMahSarlos11} and
Drineas et al.~\cite{DMMW12_JMLR},
as bounded in Eqn.~(\ref{eqn:ErrorWCE}).

Writing $Y$ as $X \beta + \epsilon$, where $X^T \epsilon = 0$, the worst-case error can be re-expressed as:
\begin{equation*}
C_{WC}(S) = \sup_{Y= X \beta + \epsilon,\; X^T \epsilon = 0} \frac{\|Y - X \beta_S\|_2^2}{\|Y - X \beta_{OLS}\|_2^2}.
\end{equation*}
Hence, in the worst-case algorithmic setup, we take a supremum over $\epsilon$, where $X^T \epsilon = 0$, whereas in the statistical setup, we take an expectation over $\epsilon$ where $\mathbb{E}[\epsilon] = 0$.

Before proceeding, several other comments about this algorithmic-statistical 
framework and our objectives are worth mentioning.
\begin{itemize}
\item
From the perspective of our two linear models, we have that 
$\beta_{OLS} = \beta + (X^T X)^{-1} X^T \epsilon$. 
In the statistical setting, since 
$\mathbb{E}[\epsilon \epsilon^T] = I_{n \times n}$, it follows that 
$\beta_{OLS}$ is a random variable with $\mathbb{E}[\beta_{OLS}] = \beta$ 
and $\mathbb{E}[(\beta - \beta_{OLS})(\beta - \beta_{OLS})^T] = (X^T X)^{-1}$.
In the algorithmic setting, on the other hand, since $X^T \epsilon = 0$, it follows that $\beta_{OLS} = \beta$. 
\item
$C_{RE}(S)$ is a statistical analogue of the worst-case algorithmic 
objective $C_{WC}(S)$, since both consider the ratio of the metrics 
$\frac{\|Y - X \beta_S\|_2^2}{\|Y - X \beta_{OLS}\|_2^2}$.
The difference is that a $\sup$ over $Y$ in the algorithmic setting is 
replaced by an expectation over noise $\epsilon$ in the statistical setting. 
A natural question is whether there is an algorithmic analogue of $C_{PE}(S)$. 
Such a performance metric would be:
\begin{equation}
\label{eqn:nonexistent_obj}
\sup_{Y} \frac{\|X (\beta - \beta_S)\|_2^2}{\|X (\beta-\beta_{OLS})\|_2^2},
\end{equation}
where $\beta$ is the projection of $Y$ on to the range space of $X^T$.  
However, since $\beta_{OLS} = \beta + (X^T X)^{-1} X^T \epsilon$ and since $X^T \epsilon = 0$, $\beta_{OLS} = \beta $ in the 
algorithmic setting, the denominator 
of Eqn.~(\ref{eqn:nonexistent_obj}) equals zero, and thus the objective in 
Eqn.~(\ref{eqn:nonexistent_obj}) is not well-defined.
The ``difficulty'' of computing or approximating this objective parallels 
our results below that show that approximating $C_{PE}(S)$ is much more 
challenging (in terms of the number of samples needed) than approximating 
$C_{RE}(S)$. 
\item
In the algorithmic setting, the sketching matrix $S$ and the objective
$C_{WC}(S)$ can depend on $X$ and $Y$ in any arbitrary way, but in the 
following we consider only sketching matrices that are either independent 
of both $X$ and $Y$ or depend only on $X$ (e.g., via the statistical 
leverage scores of $X$). 
In the statistical setting, $S$ is allowed to depend on $X$, but not on 
$Y$, as any dependence of $S$ on $Y$ might introduce correlation between 
the sketching matrix and the noise variable $\epsilon$.
Removing this restriction is of interest, especially in light of the recent 
results that show that one can obtain WC 
bounds of the form Eqn.~(\ref{eqn:ErrorWCE}) by constructing $S$ by randomly 
sampling according to an importance sampling distribution that depends on 
the \emph{influence scores}---essentially the leverage scores of the matrix $X$ 
augmented with $-Y$ as an additional column---of the $(X, Y)$ pair.
\item
Both $C_{PE}(S)$ and $C_{RE}(S)$ are qualitatively related to quantities 
analyzed by Ma et al.~\cite{MMY15_JMLR}. 
In addition, $C_{WC}(S)$ is qualitatively similar to 
$\mbox{Cov}(\widehat{\beta} | Y)$ in Ma et al., since in the algorithmic setting $Y$ is 
treated as fixed; and $C_{RE}(S)$ is qualitatively similar to 
$\mbox{Cov}(\widehat{\beta})$ in Ma et al., since in the statistical setting $Y$ is 
treated as random and coming from a linear model. 
That being said, the metrics and results we present in this paper are not 
directly comparable to those of Ma et al. since, e.g., they had a slightly different setup 
than we have here, and since they used a first-order Taylor approximation while we 
do not.
\end{itemize}

\subsection{Structural results on sketching matrices}

We are now ready to develop structural conditions characterizing how the 
sketching matrix $S$ interacts with the data $X$; this will allow us 
to provide upper bounds for the quantities $C_{WC}(S), C_{PE}(S)$, and 
$C_{RE}(S)$. 
To do this, recall that given the data matrix $X$, we can express the 
singular value decomposition of $X$ as $X = U \Sigma V^T$, where 
$U \in \mathbb{R}^{n \times p}$ is an orthogonal matrix, i.e., 
$U^T U  = I_{p \times p}$. 
In addition, we can define the \emph{oblique projection}~matrix
\begin{equation}
\Pi_S^U := U (SU)^\dagger S  .
\end{equation} 
Note that if $\mbox{rank}(SX) = p$, then $\Pi_S^U$ can be expressed as
$\Pi_S^U = U (U^T S^T S U)^{-1} U^T S^T S$, since $U^T S^T S U$ is 
invertible.
Importantly however, depending on the properties of $X$ and how $S$ is 
constructed, it can easily happen that $\mbox{rank}(SX) < p$, even if 
$\mbox{rank}(X) = p$. 

Given this setup, we can now state the following lemma, which 
characterizes how $C_{WC}(S)$, $C_{PE}(S)$, and $C_{RE}(S)$ 
depend on different structural properties of $\Pi_S^U$ and $SU$. 

\blems
\label{LemProj}
For the algorithmic setting, 
\begin{eqnarray*}
C_{WC}(S) &=& 1 + \\
& & 
\hspace{-20mm}
\sup_{\delta \in \mathbb{R}^p, U^T \epsilon = 0 } \biggr[ \frac{\| (I_{p \times p} - (SU)^{\dagger}(SU) )\delta\|_2^2}{\|\epsilon\|_2^2} + \frac{\|\Pi_S^U \epsilon \|_2^2}{\|\epsilon\|_2^2}\biggr].
\end{eqnarray*}
For the statistical setting,
\begin{equation*}
C_{PE}(S) = \frac{\| (I
_{p \times p}- (SU)^{\dagger} SU) \Sigma V^T \beta\|_2^2}{p} + \frac{\|\Pi_S^U\|_F^2}{p},
\end{equation*}
and
\begin{equation*}
C_{RE}(S) = 1+ \frac{C_{SPE}(S) - 1}{n/p - 1 } .
\end{equation*}
\elems

\noindent
Several points are worth making about Lemma~\ref{LemProj}.
\begin{itemize}
\item
For all $3$ criteria, the term which involves $(SU)^{\dagger} SU$ is a 
``bias'' term that is non-zero in the case that $\mbox{rank}(SU) < p$. 
For $C_{PE}(S)$ and $C_{RE}(S)$, the term corresponds exactly to the 
statistical bias; and if $\mbox{rank}(SU) = p$, meaning that $S$ is a 
\emph{rank-preserving} sketching matrix, then the bias term equals $0$, 
since $(SU)^{\dagger} SU = I_{p \times p}$. 
In practice, if $r$ is chosen smaller than $p$ or larger than but very close 
to $p$, it may happen that $\mbox{rank}(SU) < p$, in which case this bias is 
incurred.
\item
The final equality $C_{RE}(S) = 1+ \frac{C_{PE}(S) - 1}{n/p - 1 }$ shows 
that in general it is much more difficult (in terms of the number of 
samples needed) to obtain bounds on $C_{PE}(S)$ than $C_{RE}(S)$---since 
$C_{RE}(S)$ re-scales $C_{PE}(S)$ by $p/n$, which is much less than $1$. 
This will be reflected in the main results below, where the scaling of 
$C_{RE}(S)$ will be a factor of $p/n$ smaller than $C_{PE}(S)$.  
In general, it is significantly more difficult to bound $C_{PE}(S)$, since 
$\|X(\beta - \beta_{OLS})\|_2^2$ is $p$, whereas $\|Y - X \beta_{OLS}\|_2^2$ 
is $n-p$, and so there is much less margin for error in approximating 
$C_{PE}(S)$. 
\item
In the algorithmic or worst-case setting, 
$\sup_{\epsilon \in \mathbb{R}^n/\{ 0\}, \Pi^U \epsilon = 0 } \frac{\|\Pi_S^U \epsilon \|_2^2}{\|\epsilon\|_2^2}$ 
is the relevant quantity, whereas in the statistical setting 
$\|\Pi_S^U\|_F^2$ is the relevant quantity. 
The Frobenius norm enters in the statistical setting because we are taking 
an average over homoscedastic noise, and so the $\ell_2$ norm of the 
eigenvalues of $\Pi_S^U$ need to be controlled. 
On the other hand, in the algorithmic or worst-case setting, the worst 
direction in the null-space of $U^T$ needs to be controlled, and thus the 
spectral norm enters.
\end{itemize}

\section{Main theoretical results}
\label{SecMainResults}

In this section, we provide upper bounds for $C_{WC}(S)$, $C_{PE}(S)$, 
and $C_{RE}(S)$, where $S$ correspond to random sampling and random 
projection matrices. 
In particular, we provide upper bounds for $4$ sketching matrices:
(1) a vanilla leverage-based random sampling algorithm from 
Drineas et al.~\cite{DMMW12_JMLR}; 
(2) a variant of leverage-based random sampling, where the random samples
are \emph{not} re-scaled prior to their inclusion in the sketch;
(3) a vanilla random projection algorithm, where $S$ is a random matrix 
containing i.i.d. sub-Gaussian random variables; and 
(4) a random projection algorithm, where $S$ is a random 
Hadamard-based random projection, as analyzed in~\cite{BoutsGitt12}.

\subsection{Random sampling methods}
\label{SecSampling}

Here, we consider random sampling algorithms.
To do so, first define a random sampling matrix $\tilde{S} \in \mathbb{R}^n$ 
as follows: $\tilde{S}_{ij} \in \{0, 1\}$ for all $(i,j)$ and 
$\sum_{j=1}^n \tilde{S}_{ij} = 1$, where each row has an independent 
multinomial distribution with probabilities $(p_i)_{i=1}^n$. 
The matrix of cross-leverage scores is defined as 
$L = U U^T \in \mathbb{R}^{n \times n}$, and $\ell_i = L_{ii}$ denotes the 
leverage score corresponding to the $i^{th}$ sample. 
Note that the leverage scores satisfy 
$\sum_{i=1}^n{\ell_i} = \mbox{trace}(L) = p$ and $0 \leq \ell_i \leq 1$.
 
The sampling probability distribution we consider $(p_i)_{i=1}^n$ is of the 
form $p_i = (1 - \theta) \frac{\ell_i}{p} + \theta q_i$, where 
$\{q_i\}_{i=1}^n$ satisfies $0 \leq q_i \leq 1$ and $\sum_{i=1}^n {q_i} = 1$ is 
an arbitrary probability distribution, and $0 \leq \theta < 1$. In other words, $p_i$ is a convex combination of a leverage-based distribution and another arbitrary distribution. Note that for $\theta = 0$, the probabilities are proportional to the leverage scores, whereas for $\theta = 1$, the probabilities follow the distribution defined by $\{q_i\}_{i=1}^n$.

We consider two sampling matrices, one where the random sampling matrix is 
re-scaled, as in Drineas et al.~\cite{DrinMuthuMahSarlos11}, and one in 
which no re-scaling takes place. In particular, let $S_{NR} = \tilde{S}$ denote the random sampling matrix (where the subscript $NR$ denotes the fact that no re-scaling takes place). The re-scaled sampling matrix is 
$S_{R} \in \mathbb{R}^{r \times n} = \tilde{S} W$, where 
$W \in \mathbb{R}^{n \times n}$ is a diagonal re-scaling matrix, where 
$[W]_{jj} = \sqrt{\frac{1}{r p_j}}$ and $W_{ji} = 0$ for $j \neq i$. 
The quantity $\frac{1}{p_j}$ is the re-scaling factor.

In this case, we have the following result.

\btheos
\label{ThmOne}
For $S = S_{R}$, with $r \geq \frac{C p}{(1-\theta)} \log\big(\frac{C' p}{(1-\theta)} \big)$,
then with probability at least $0.7$,
it holds that $\mbox{rank}(S_R U) = p$ and that:
\begin{eqnarray*}
C_{WC}(S_{R}) & \leq & 1+12 \frac{p}{r} \\
C_{PE}(S_{R}) & \leq & 44 \frac{n}{r}\\
C_{RE}(S_{R}) & \leq & 1+ 44 \frac{p}{r}  .
\end{eqnarray*}
\etheos
\vspace{-3mm}
Several things are worth noting about this result.
First, note that both $C_{WC}(S_{R})-1$ and $C_{RE}(S_{R})-1$ scale as 
$\frac{p}{r}$; thus, it is possible to obtain high-quality performance 
guarantees for ordinary least squares, as long as $\frac{p}{r} \rightarrow 0$, 
e.g., if $r$ is only slightly larger than $p$.
On the other hand, $C_{PE}(S_{R})$ scales as $\frac{n}{r}$, meaning $r$ 
needs to be close to $n$ to provide similar performance guarantees. 
Next, note that all of the upper bounds apply to any data matrix $X$, 
without assuming any additional structure on $X$. 

Also note that the distribution $\{q_i\}_{i=1}^n$ does not enter the results which means our bounds hold for any choice of $\{q_i\}_{i=1}^n$ and don't depend on $\theta$. This allows to consider different distributions. A standard choice is uniform, i.e., $q_i = \frac{1}{n}$ (see e.g. Ma et al.~\cite{MMY15_JMLR}). The other important example is that of \emph{approximate} leverage-score sampling developed in ~\cite{DMMW12_JMLR} that reduces computation.  Let $(\tilde{\ell_i})_{i=1}^n$ denote the approximate leverage scores developed by the procedure in~\cite{DMMW12_JMLR}. Based on Theorem 2 in ~\cite{DMMW12_JMLR}, $|\ell_i - \tilde{\ell_i}| \leq \theta$ where $0 < \theta < 1$ for $r$ sufficiently large. Now, using $p_i = \frac{\tilde{\ell_i}}{p}$, $p_i$ can be re-expressed as $p_i = (1-\theta) \frac{\ell_i}{p} + \theta q_i$ where $(q_i)_{i=1}^n$ is a distribution (unknown since we only have a bound on the approximate leverage scores). Hence, the performance bounds achieved by approximate leveraging are equivalent to those achieved by adding $\theta$ multiplied by a uniform or other arbitrary distribution. 

Next, we consider the leverage-score estimator without re-scaling $S_{NR}$. In order to develop nontrivial bounds on $C_{WC}(S_{NR})$, $C_{PE}(S_{NR})$, and $C_{RE}(S_{NR})$, we need to make a (strong) assumption on the leverage-score distribution on $X$.
To do so, we define the following.

\bdes[k-heavy hitter leverage distribution]
A sequence of leverage scores $(\ell_i)_{i=1}^n$ is a \emph{k-heavy hitter} leverage score distribution if there exist constants $c, C > 0$ such that for $1 \leq i \leq k$, $\frac{c p}{k} \leq \ell_i \leq \frac{C p}{k}$ and for the remaining $n-k$ leverage scores, $\sum_{i=k+1}^p {\ell_i} \leq \frac{3}{4}$.
\edes
\noindent
The interpretation of a $k$-heavy hitter leverage distribution is one in which only $k$ samples in $X$ contain the majority of the leverage score mass. The parameter $k$ acts as a measure of non-uniformity, in that the smaller the $k$, the more non-uniform are the leverage scores. The $k$-heavy hitter leverage distribution allows us to model highly non-uniform leverage scores which allows us to state the following result.

\btheos
\label{ThmTwo}
For $S = S_{NR}$, with $\theta = 0$ and assuming a $k$-heavy hitter leverage distribution and $r \geq c_1 p \log\big(c_2 p\big)$, 
then with probability at least $0.6$,
it holds that $\mbox{rank}(S_{NR}) = p$ and that:
\begin{eqnarray*}
C_{WC}(S_{NR}) & \leq & 1+ \frac{44 C^2}{c^2} \frac{p}{r} \\
C_{PE}(S_{NR}) & \leq & \frac{44 C^4}{c^2} \frac{k}{r}\\
C_{RE}(S_{NR}) & \leq & 1 + \frac{44 C^4}{c^2} \frac{p k}{n r}  .
\end{eqnarray*}
\etheos
\vspace{-3mm}
Notice that when $k \ll n$, bounds in Theorem~\ref{ThmTwo} on $C_{PE}(S_{NR})$ and $C_{RE}(S_{NR})$ are significantly sharper than bounds in Theorem~\ref{ThmOne} on $C_{PE}(S_{R})$ and $C_{RE}(S_{R})$. Hence not re-scaling has the potential to provide sharper bound in the statistical setting. However a much stronger assumption on $X$ is needed for this~result.

\vspace{-3mm}
\subsection{Random projection methods}

Here, we consider two random projection algorithms, one based on a sub-Gaussian projection matrix and the other based on a Hadamard projection matrix. To do so, define $[S_{SGP}]_{ij} = \frac{1}{\sqrt{r}} X_{ij}$, where $(X_{ij})_{1\leq i \leq r, 1 \leq j \leq n}$ are i.i.d. sub-Gaussian random variables with $\mathbb{E}[X_{ij}] = 0$, variance $\mathbb{E}[X_{ij}^2] = \sigma^2$ and sub-Gaussian parameter $1$. 
In this case, we have the following result.

\btheos
\label{ThmThree}
For any matrix $X$, there exists a constant $c$ such that if $r \geq c' \log n$, 
then with probability greater than $0.7$, 
it holds that $\mbox{rank}(S_{SGP}) = p$ and that:
\begin{eqnarray*}
C_{WC}(S_{SGP}) & \leq & 1 + 11 \frac{p}{r}\\
C_{PE}(S_{SGP}) & \leq & 44(1 + \frac{n}{r})\\
C_{RE}(S_{SGP}) & \leq & 1 + 44 \frac{p}{r}  .
\end{eqnarray*}
\etheos
\vspace{-3mm}
Notice that the bounds in Theorem~\ref{ThmThree} for $S_{SGP}$ are equivalent to the bounds in Theorem~\ref{ThmOne} for $S_{R}$, except that $r$ is required only to be larger than $O(\log n)$ rather than $O(p \log p)$. Hence for smaller values of $p$, random sub-Gaussian projections are more stable than leverage-score sampling based approaches. This reflects the fact that to a first-order approximation, leverage-score sampling performs as well as performing a smooth projection. 

Next, we consider the randomized Hadamard projection matrix. In particular, $S_{Had} = S_{Unif} H D$, where $H \in \mathbb{R}^{n \times n}$ is the standard Hadamard matrix (see e.g.~\cite{Hedayat78}), $S_{Unif} \in \mathbb{R}^{r \times n}$ is an $r \times n$ uniform sampling matrix, and $D \in \mathbb{R}^{n \times n}$ is a diagonal matrix with random equiprobable $\pm 1$ entries.

\btheos
\label{ThmFour}
For any matrix $X$, there exists a constant $c$ such that if $r \geq c p \log n ( \log p + \log \log n)$, 
then with probability greater than $0.8$, 
it holds that $\mbox{rank}(S_{Had}) = p$ and that:
\begin{eqnarray*}
C_{WC}(S_{Had}) & \leq & 1 + 40 \log(np) \frac{p}{r}\\
C_{RE}(S_{Had}) & \leq & 40\log (np) (1 + \frac{n}{r})\\
C_{PE}(S_{Had}) & \leq & 1 + 40\log (np) (1 + \frac{p}{r}).  
\end{eqnarray*}
\etheos
\vspace{-3mm}
Notice that the bounds in Theorem~\ref{ThmFour} for $S_{Had}$ are equivalent to the bounds in Theorem~\ref{ThmOne} for $S_{R}$, up to a constant and $\log(np)$ factor. As discussed in Drineas et al.~\cite{DrinMuthuMahSarlos11}, the Hadamard transformation makes the leverage scores of $X$ approximately uniform (up to $\log(np)$ factor), which is why the performance is similar to the sub-Gaussian projection (which also tends to make the leverage scores of $X$ approximately uniform). We suspect that the additional $\log(np)$ factor is an artifact of the analysis since we use an entry-wise concentration bound; using more sophisticated techniques, we believe that the $\log(np)$ can be improved.

\vspace{-3mm}
\subsection{Lower Bounds}
\label{SecLower}

In concurrent work, Pilanci and Wainwright~\cite{PilanciWainwright} amongst other results develop lower bounds on the numerator in $C_{PE}(S)$ which prove that our upper bounds on $C_{PE}(S)$ can not be improved. We re-state Theorem 1 (Example 1) in Pilanci and Wainwright~\cite{PilanciWainwright} in a way that makes it most comparable to our results.

\btheos[Theorem 1 in ~\cite{PilanciWainwright}]
For any sketching matrix satisfying $\|\mathbb{E}[S^T(S S^T)^{-1}S]\|_{op} \leq \eta \frac{r}{n}$, any estimator based on $(SX, SY)$ satisfies the lower bound with probability greater than $1/2$:
\begin{eqnarray*}
C_{PE}(S) & \geq & \frac{n}{128 \eta r}.
\end{eqnarray*}
\etheos

Gaussian and Hadamard projections as well as re-weighted approximate leverage-score sampling all satisfy the condition $\|\mathbb{E}[S^T(S S^T)^{-1}S]\|_{op} \leq \eta \frac{r}{n}$. On the other hand un-weighted leverage-score sampling does not satisfy this condition and hence does not satisfy the lower bound which is why we are able to prove a tighter upper bound when the matrix $X$ has highly non-uniform leverage scores. This proves that $C_{PE}(S)$ is a quantity that is more challenging to control than $C_{RE}(S)$ and $C_{WC}(S)$ when only a single sketch is used. Using this insight, Pilanci and Wainwright~\cite{PilanciWainwright} show that by using a particular iterative Hessian sketch, $C_{PE}(S)$ can be controlled up to constant.  In addition to providing a lower bound on the PE using a sketching matrix just once, Pilanci and Wainwright also develop a new iterative sketcthing scheme where sketching matrices are used repeatedly can reduce the PE significantly.

\vspace{-3mm}
\section{Discussion and conclusion}
\label{SecDiscussion}

In this paper, we developed a framework for analyzing algorithmic and statistical criteria for general sketching matrices $S \in \mathbb{R}^{r \times n}$ applied to the LS objective. Our framework reveals that the algorithmic and statistical criteria depend on different properties of the oblique projection matrix $\Pi^U_S = U(SU)^{\dagger} U$, where $U$ is the left singular matrix for $X$. In particular, the algorithmic WC criteria depends on the quantity $\sup_{U^T \epsilon = 0} \frac{\|\Pi^U_S \epsilon \|_2}{\|\epsilon\|_2}$, since in that case the data may be arbitrary and worst-case, whereas the two statistical criteria (RE and PE) depends on $\| \Pi^U_S\|_F$, since in that case the data follow a linear model with homogenous noise variance.

Using our framework we develop upper bounds for three performance criterion applied to $4$ sketching schemes. Our upper bounds reveal that in the regime where $ p < r \ll n$, our sketching schemes achieve optimal performance up to constant in terms of WC and RE. On the other hand, the PE scales as $\frac{n}{r}$ meaning $r$ needs to be close to $n$ for good performance; and subsequent lower bounds in Pilanci and Wainwright~\cite{PilanciWainwright} show that this upper bound can not be improved.

\bibliography{PaperFinal}
\bibliographystyle{icml2015}

\end{document}